\documentclass{article}
\usepackage{microtype}
\usepackage{graphicx}
\usepackage{subfigure}
\usepackage{booktabs} % for professional tables
\usepackage{array,booktabs,arydshln,xcolor}
\usepackage{graphicx,wrapfig,lipsum}
\usepackage{float}
\usepackage{natbib}

\usepackage{hyperref}
\usepackage{verbatim}
\usepackage{amssymb}
\usepackage{array}

\usepackage[ruled,linesnumbered,vlined]{algorithm2e}
\newcommand{\PreserveBackslash}[1]{\let\temp=\\#1\let\\=\temp}
\newcolumntype{C}[1]{>{\PreserveBackslash\centering}p{#1}}

\usepackage{amsmath}
\usepackage[noabbrev,capitalize]{cleveref}
\usepackage{bbm, dsfont}
\usepackage[preprint]{neurips_2020}
\usepackage{nccmath}
\usepackage{mathtools}
\usepackage{nicefrac} 
\usepackage{xfrac} 

\usepackage{lipsum}

\usepackage{multicol}
\usepackage{lipsum}
\usepackage{tikz}
\usetikzlibrary{positioning,angles,quotes,arrows,automata}

\usepackage{float}
\usepackage{bm}
\usepackage{amsfonts}
\usepackage{amsbsy}
\usepackage{amsthm}
\usepackage{color}
\usepackage{amsmath}
\usepackage{enumerate}% http://ctan.org/pkg/enumerate
\usepackage{amssymb}
\usepackage{enumitem}
\usepackage{array}
\usepackage{nicefrac}

\usepackage{tabularx}
\usepackage{booktabs}
\usepackage{multirow}
\newcolumntype{Y}{>{\centering\arraybackslash}X}

\usepackage{color, colortbl}
\definecolor{Gray}{gray}{0.98}
\definecolor{LightCyan}{rgb}{0.88,1,1}
\newcolumntype{g}{>{\columncolor{Gray}}c}

\DeclareMathOperator*{\argmax}{argmax}
\DeclarePairedDelimiter\abs{\lvert}{\rvert}%
\DeclarePairedDelimiter\norm{\lVert}{\rVert}%
\makeatletter
\let\oldabs\abs
\def\abs{\@ifstar{\oldabs}{\oldabs*}}
\let\oldnorm\norm
\def\norm{\@ifstar{\oldnorm}{\oldnorm*}}
\makeatother

\newcommand{\eat}[1]{}
\newcommand{\states}{\mathcal{S}}
\newcommand{\actions}{\mathcal{A}}

\newcommand*{\tr}{^{\mkern-1.5mu\mathsf{T}}}

\newcommand{\cs}{\\[1ex] & }

\newcommand{\Real}{\mathbb{R}}
\renewcommand{\ss}{\,:\,}

\newcommand{\simplexs}{\Delta ^{ \states }}

\newcommand{\ambset}{\mathcal{P}}

\newcommand{\dataset}{\mathcal{D}}
\newcommand{\opt}{^\star}

\usepackage{thmtools}
\usepackage{nameref}
\usepackage[noabbrev,capitalize]{cleveref}

\makeatletter
\newcommand{\myref}[1]{\cref{#1}\mynameref{#1}{\csname r@#1\endcsname}}
\newcommand{\Myref}[1]{\Cref{#1}\mynameref{#1}{\csname r@#1\endcsname}}

\def\mynameref#1#2{%
	\begingroup
	\edef\@mytxt{#2}%
	\edef\@mytst{\expandafter\@thirdoffive\@mytxt}%
	\ifx\@mytst\empty\else
	\space(\nameref{#1})\fi
	\endgroup
}
\makeatother

\DeclareMathOperator{\E}{\mathbb{E}}

\renewcommand{\ss}{\,:\,}
\newcommand{\statecount}{S}

\newcommand{\RBU}{\widehat{T}}

\title{Robust Constrained-MDPs: Soft-Constrained Robust Policy Optimization under Model Uncertainty}
\author{
Reazul Hasan Russel\hspace{16pt} Mouhacine Benosman \hspace{16pt} Jeroen Van Baar\\ 
Mitsubishi
Electric Research Laboratories (MERL)\\ Cambridge, MA 02139, USA\\
\{{\tt rrussel}, {\tt benosman}, {\tt jeoren}\}$@$ {\tt merl.com}
}

\date{}

\begin{document}
\maketitle

\begin{abstract}
In this paper, we focus on the problem of robustifying reinforcement learning (RL) algorithms with respect to model uncertainties. Indeed, in the framework of model-based RL, we propose to merge the theory of constrained Markov decision process (CMDP), with the theory of robust Markov decision process (RMDP), leading to a formulation of robust constrained-MDPs (RCMDP). This formulation, simple in essence, allows us to design RL algorithms that are robust in performance, and provides constraint satisfaction guarantees, with respect to uncertainties in the system's states transition probabilities. The need for RCMPDs is important for real-life applications of RL. For instance, such formulation can play an important role for policy transfer from simulation to real world (Sim2Real) in safety critical applications, which would benefit from performance and safety guarantees which are robust w.r.t model uncertainty. We first propose the general problem formulation under the concept of RCMDP, and then propose a Lagrangian formulation of the optimal problem, leading to a robust-constrained policy gradient RL algorithm. We finally validate this concept on the inventory management problem.  
\end{abstract}

\section{Introduction}
Reinforcement learning (RL) is a learning framework that addresses sequential decision-making problems, wherein an `agent' or a decision maker learns a policy to optimize a long-term reward by interacting with the (unknown or partially known) environment. At each step, the RL agent obtains evaluative feedback (called reward or cost) about the performance of its action, allowing it to improve the performance of subsequent actions \citep{sutton1998}. With the advent of deep learning, RL has witnessed huge successes in recent times~\citep{silver2017mastering}. However, since most of these methods rely on model-free RL, there are several unsolved challenges, which restrict the use of these algorithms for many safety critical physical systems~\citep{vamtutorial,Benosman2018}. For example, it is very difficult for most model-free RL algorithms to ensure basic properties like stability of solutions, robustness with respect to model uncertainties, etc. This has led to several research directions which study incorporating robustness, constraint satisfaction, and safe exploration during learning for safety critical applications. While safe exploration and robust stability guarantees are highly desirable, they are also very challenging to incorporate in RL algorithms. The main goal of our work is to formulate this incorporation into robust constrained-MDPs (RCMDPs), and derive the corresponding equations necessary to solve problems on RCMDPs.

Constrained Markov decision processes (CMDPs) can be seen as an extension of MDPs with expected cumulative cost constraints, e.g.,~\citep{Altman2004}. For such CMDPs, several solution methods have been proposed, e.g., linear programming-based solutions~\citep{Altman2004}, surrogate-based methods~\citep{CYA16,Dalal2018}, Lagrangian methods~\citep{Geibel2005,Altman2004}. We refer to these CMDPS as {\it non-robust}  since they do not take uncertainties in the state transition probability into account, which is an important factor in real-life applications.

%The existing formulations of CMDPs are specific to the case of known models, without uncertainties. 
%However, all these methods do not take into account uncertainties in the state transition probability, which is an important factor in real-life applications.

On the other hand, robustification of MDPs, w.r.t. model uncertainties, can be found in the context of robust MDPs (RMDPs) which generalize MDPs to the case where transition probabilities and/or rewards are not perfectly known, e.g.,~\cite{Nilim2004,Wiesemann2013}. These RMDPs can be formulated and solved using so-called ambiguity or uncertainty sets, e.g.,~\citep{petrik2019,petrik2012,petrik2016}. However, one noticeable point in all the RMDPs-based RL algorithms is the fact that they do not consider any safety constraints, i.e., expected cumulative cost constraints. 

These safety constraints are important in real-life applications, where one cannot afford to risk violating some given constraints, e.g., in autonomous cars, there are hard safety constraints on the robot velocities and steering angles. Besides, in real applications, to mitigate the sample inefficiency of model-free RL algorithms, training often occurs on a simulated environment. The result is then transferred to the real world, typically followed by fine-tuning, a process referred to as Sim2Real. The simulation model is by definition uncertain with respect to the real world, due to approximations and lack of system identification. Domain randomization \citep{vanBaar2019may} and meta-learning \citep{pmlr-v78-finn17a}, aimed at addressing model uncertainty in transfer, offer no guarantees. Furthermore, for safety critical applications, a trained policy in simulation should offer certain guarantees on safety when transferred to the real world.
 
 In light of these practical motivations, we propose to merge the two concepts of CMDPs and RMDPs, to ensure both safety and robustness. In this RCMDP concept, we propose to robustify both the performance cost minimization, as well as the safety constraints (via cumulative constraint costs). Indeed, robustness is equally (if not more) important in estimating the cumulative constraint costs along the whole trajectory in order to certify that there will be no unexpected violations, if the system is deployed in reality. That is, if deployed, the worst-case cumulative constrained-cost will not exceed a pre-determined safety budget.

The contribution of this paper is four-fold: 1) Intuited from the concepts of CMDP and RMDP, we formulate the concept of RCMDP, bridging the gap between constraints and robustness w.r.t. state transition probability uncertainties; 2) propose a robust soft-constrained Lagrange-based solution of the RCMDP problem; 3) derive associated gradient update rule and present a policy gradient algorithm;  4) illustrate the performance of the proposed algorithm on the inventory management problem under model uncertainties.

The paper is organized as follows: \cref{sec:formulation} describes the formulation of our Robust-CMDP problem and the objective we seek to optimize. A Lagrange based approach is presented in \cref{sec:rc_opt} along with required gradient update rules and a robust constrained policy gradient algorithm. We evaluate our algorithm in \cref{sec:experiment} and draw the concluding remarks in \cref{sec:conclusion}.

\section{Problem Formulation} \label{sec:formulation}
We consider a robust-MDP model with a finite number of states $\states = \{1, \ldots, S \}$ and finite number of actions $\actions = \{1, \ldots, A\}$. Every action $a \in \actions$ is available for the decision maker to take in every state $s \in \states$. After taking an action $a\in\actions$ in state $s\in\states$, the decision maker receives a cost $c(s,a) \in \Real$ and transitions to a next state $s'$ according to the \emph{true} but \emph{unknown} transition probability $p_{s,a} \opt \in \simplexs$. $p_0$ is the distribution of an initial state. We further incorporate a constrained-MDP~\citep{Altman2004} setup into this robust-MDP model by introducing a constraint cost $d(s)\in [0,D_{\max} ]$ and an associated constraint budget $d_0\in\Real_+$.

An ambiguity set $\ambset_{s,a}$, defined for each state $s\in\states$ and action $a\in\actions$, is a set of feasible transition matrices quantifying the uncertainty in transition probabilities. In this paper, we restrict our attention to $s,a-$rectangular ambiguity sets which simply assumes independence between different state-action pairs~\citep{LeTallec2007,Wiesemann2013}.

We use in this paper $L_1-$norm bounded ambiguity sets around the nominal transition probability $\bar{p}_{s,a}=\E[p\opt_{s,a}|\dataset]$, on some dataset \(\dataset\), as:
\begin{equation*}
\ambset_{s,a} = \bigl\{p \in \Delta^\statecount \ss \norm{p - \bar{p}_{s,a} }_1 \le \psi_{s,a} \bigr\}
\end{equation*}
Where $\psi_{s,a}\ge 0$ is the budget of allowed deviations. This budget $\psi$ can be computed using Hoeffding bound as~\citep{petrik2019beyond}: $\psi_{s,a} = \sqrt{\frac{2}{n_{s,a}} \log \frac{S A 2^{S}}{\delta} }$, where $n_{s,a}$ is the number of transitions in dataset $\dataset$ originating from state $s$ and an action $a$, and $\delta$ is the confidence level. Note that this is just one specific choice for the ambiguity set. Our method can be extended to any other type of ambiguity sets (e.g. $L_\infty-$norm, Bayesian, weighted etc.). We use $\ambset$ to refer cumulatively to $\ambset_{s,a}$ for all states $s\in\states$ and actions $a\in\actions$.  

A stationary randomized policy $\pi(\cdot|s)$ for state $s\in\states$ defines a probability distribution over actions $a\in\actions$, $\Pi$ represents the set of all stationary randomized policies. We parameterize the randomized policy for state $s\in\states$ as $\pi(\cdot|s;\theta)$ where $\theta \subseteq \Real^k$ is a $k-$dimensional parameter vector. Let $\xi=\{s_0,a_0,c_0,d_0,\ldots,s_{T-1},a_{T-1},c_{T-1},d_{T-1},s_T\}$ be a sampled trajectory generated by executing a policy $\pi$ from a starting state $s_0$. Then the probability of sampling $\xi$ is: $p^\theta(\xi)=p_0(s_0)\prod_{t=0}^{T-1}\pi(a_t|s_t;\theta)p(s_{t+1}|s_t,a_t)$. The total cost $g$ for trajectory $\xi$ is: $g(\xi) =  \sum_{t=0}^{\infty} \gamma^t c(s_t,a_t)$~\citep{Puterman2005}. The value function is defined as the expected return: $v^\theta(s_0) = \E_{p}\big[ g(\xi) \big]$. We define the robust value function $\hat{v}$ as the expected return in the worst-case realization of the transition probability within $\ambset$ as: $\hat{v}^\pi_\ambset(s_0) = \max_{p\in\ambset}\E_p\big[ g(\xi)\big]$. Similarly, the total constraint-cost for trajectory $\xi$ is: $h(\xi) =  \sum_{t=0}^{\infty} \gamma^t d(s_t,a_t)$. And the robust constraint value function $\hat{u}$ is defined as: $\hat{u}^\pi_\ambset(s_0) = \max_{p\in\ambset}\E_p\big[ h(\xi)\big]$.

The robust Bellman operator $\RBU_\ambset$ for a state $s\in\states$ and an ambiguity set $\ambset$ computes the best action with respect to the worst-case realization of the transition probabilities in $\ambset$ as:
\begin{equation*} \label{eq:bellman_definition}
\begin{aligned}
    (\RBU_\ambset v)(s) := \min_{a\in\actions}\max_{p \in\ambset_{s,a}}  (c(s,a) + \gamma \cdot p\tr v)
\end{aligned}
\end{equation*}
The optimal robust value function $\hat{v}\opt$, and the robust value function $\hat{v}^\pi$ for a policy $\pi$ are unique and satisfy $\hat{v}\opt = \RBU_\ambset \hat{v}\opt$ and  $\hat{v}^\pi = \RBU_\ambset^\pi \hat{v}^\pi$ ~\citep{Iyengar2005}. Similarly, all these properties hold for the constrained robust value function $\hat{u}$ as well.

\paragraph{Objective} Our objective is then to solve the RCMDP optimization problem below:
\begin{equation} \label{eq:rcmdp_obj}
    \begin{aligned}
      &\min_{\pi\in\Pi} \hat{v}^\pi_\ambset(s)\\
      &\text{s.t. } \hat{u}^\pi_\ambset(s) \le d_0
    \end{aligned}
\end{equation}
This objective resembles the objective of a CMDP, but with additional robustness integrated by the quantification of the uncertainty in the model.

\section{Robust Constrained Optimization} \label{sec:rc_opt}
A general approach for solving \cref{eq:rcmdp_obj} is to apply the Lagrange relaxation procedure (Chapter 3 of \cite{Bertsekas2003}), which turns it into an unconstrained optimization problem:
\begin{equation} \label{eq:crmdp_lagrange}
    \begin{aligned}
    \max_{\lambda\ge 0} \min_\theta \Bigg( L(\theta,\lambda) &= \hat{v}^\pi_\ambset(s) + \lambda \Big(\hat{u}^\pi_\ambset(s)-d_0\Big) \Bigg)
    \end{aligned}
\end{equation}
where $\lambda$ is known as the \emph{Lagrange multiplier}. The goal is then to find a saddle point $(\theta^*,\lambda^*)$ that satisfies $L(\theta,\lambda^*) \ge L(\theta^*,\lambda^*) \ge L(\theta^*,\lambda)$, $\forall \theta,  \lambda$. This is achieved by descending in $\theta$ and ascending in $\lambda$ using the gradients.

Without loss of generality, we rewrite \Cref{eq:crmdp_lagrange} for a fixed starting state $s_0$ and perform some algebraic manipulation:
\begin{equation*} \label{eq:crmdp_lagrange_extend}
    \begin{aligned}
    L(\theta,\lambda) &= \hat{v}^\pi_\ambset(s_0) + \lambda \Big(\hat{u}^\pi_\ambset(s_0)-d_0 \Big)\\
    &= \max_{p\in\ambset}\E_p\big[ g(\xi) \big] + \lambda \Big(\max_{p\in\ambset}\E_p\big[ h(\xi) \big]-d_0\Big)\\
    &\stackrel{(a)}{=} \E_{\hat{p}^\theta_v}\big[ g(\xi) \big] + \lambda \E_{\hat{p}^\theta_u} \big[ h(\xi) \big]- \lambda d_0\\
    &= \sum_\xi \bigg( \hat{p}^\theta_v(\xi) g(\xi) + \lambda \hat{p}^\theta_u(\xi) h(\xi) \bigg) - \lambda d_0
    \end{aligned}
\end{equation*}
Here $(a)$ follows with $\hat{p}^\theta_v = \argmax_{p\in\ambset}\E_p\big[ g(\xi) \big]$ and $\hat{p}^\theta_u = \argmax_{p\in\ambset}\E_p\big[ h(\xi) \big]$.
\subsection{Gradient Update Rules} \label{subsec:grad_update}
We now derive the gradient update rules with respect to $\theta$ as below:
\begin{equation*} \label{eq:grad_theta}
    \begin{aligned}
    \nabla_\theta L(\theta,\lambda) &= \sum_\xi \bigg( \nabla_\theta \hat{p}^\theta_v(\xi) g(\xi) + \lambda \nabla_\theta\hat{p}^\theta_u(\xi) h(\xi) \bigg)\\
    &= \sum_\xi \bigg( \hat{p}^\theta_v(\xi)g(\xi) \nabla_\theta \log \hat{p}^\theta_v(\xi) + \lambda \hat{p}^\theta_u(\xi)h(\xi) \nabla_\theta \log \hat{p}^\theta_u(\xi) \bigg)\\
    &= \sum_\xi \Bigg( \hat{p}^\theta_v(\xi)g(\xi) \nabla_\theta \log \bigg( p_0(s_0)\prod_{t=0}^{T-1} \hat{p}^\theta_v(s_{t+1}|s_t,a_t) \pi_\theta(a_t|s_t) \bigg) \\
    & \hspace{1.0cm}+ \lambda \hat{p}^\theta_u(\xi)h(\xi) \nabla_\theta \log \bigg( p_0(s_0)\prod_{t=0}^{T-1} \hat{p}^\theta_u(s_{t+1}|s_t,a_t) \pi_\theta(a_t|s_t) \bigg) \Bigg)\\
    &= \sum_\xi \Bigg( \hat{p}^\theta_v(\xi)g(\xi) \nabla_\theta  \bigg( \log p_0(s_0) + \sum_{t=0}^{T-1} \log \hat{p}^\theta_v(s_{t+1}|s_t,a_t) + \log \pi_\theta(a_t|s_t) \bigg) \\
    & \hspace{1.0cm}+ \lambda \hat{p}^\theta_u(\xi) h(\xi) \nabla_\theta \bigg( \log p_0(s_0) + \sum_{t=0}^{T-1} \log \hat{p}^\theta_u(s_{t+1}|s_t,a_t) + \log \pi_\theta(a_t|s_t) \bigg) \Bigg)\\
    &= \sum_\xi \bigg( \hat{p}^\theta_v(\xi) g(\xi)   \sum_{t=0}^{T-1} \nabla_\theta \log \pi_\theta(a_t|s_t)
    + \lambda \hat{p}^\theta_u(\xi) h(\xi) \sum_{t=0}^{T-1} \nabla_\theta \log \pi_\theta(a_t|s_t) \bigg)\\
    &= \sum_\xi \bigg( \hat{p}^\theta_v(\xi) g(\xi)
    + \lambda \hat{p}^\theta_u(\xi) h(\xi) \bigg) \sum_{t=0}^{T-1} \frac{\nabla_\theta \pi_\theta(a_t|s_t)}{\pi_\theta(a_t|s_t)}\\
    \end{aligned}
\end{equation*}
Notice that the constraint budget $d_0$ does not play any role in the policy optimization. Also, the expectations over the cost $g(\xi)$ and constraint cost $h(\xi)$ are with respect to $\hat{p}^\theta_v(\xi)$ and $\hat{p}^\theta_u(\xi)$), respectively. However, the costs and constraint costs are coupled together in reality, meaning that the two trajectories would not diverge. So one of $\hat{p}^\theta_v(\xi)$ or $\hat{p}^\theta_u(\xi)$ can be chosen depending on the priorities toward robustness of cost or constraint cost, and both of the expectations can be evaluated with that common probability measure $\hat{p}^\theta(\xi)$. The gradient update rule then becomes: 
\begin{equation*}
    \nabla_\theta L(\theta,\lambda) = \sum_\xi \hat{p}^\theta(\xi) \bigg( g(\xi)
    + \lambda h(\xi) \bigg) \sum_{t=0}^{T-1} \frac{\nabla_\theta \pi_\theta(a_t|s_t)}{\pi_\theta(a_t|s_t)}
\end{equation*}
And the gradient update rule with respect to $\lambda$ as below:
\begin{equation*} \label{eq:grad_lambda}
    \begin{aligned}
    \nabla_\lambda L(\theta,\lambda) &= \nabla_\lambda \Bigg( \sum_\xi \hat{p}^\theta_v(\xi) g(\xi) + \lambda \bigg( \sum_\xi \hat{p}^\theta_u(\xi) h(\xi) - d_0 \bigg) \Bigg)\\
    &= \sum_\xi \hat{p}^\theta_u(\xi) h(\xi) - d_0
    \end{aligned}
\end{equation*}

\subsection{Policy Gradient Algorithm} \label{subsec:pg_algo}
\begin{algorithm*} [!h]
	\KwIn{A differentiable policy parameterization $\pi(.|.,\theta)$, nominal transition model $\bar{p}$, step size schedules $\zeta_2$ and $\zeta_1$.}
	\KwOut{Policy parameters $\theta$}
	Initialize actor parameters $\theta\gets\theta_0$, and critic parameter $\lambda\gets \lambda_0$;
    
	\For{$k\gets0,1,2,\ldots$}{
		
	    Sample initial state: $s_0 \sim p_0$;
	    
	    Trajectory: $\xi \gets \emptyset$;
	    
	    \tcc{Simulate trajectory with current policy $\theta$}
		\For{$t\gets0,1,2,\ldots,T$}{
        Sample action: $a_t\sim \pi(\cdot|s_t,\theta)$;
        
        Observe cost: $c_t=c(s_t,a_t)$ and constraint cost: $d_t=d(s_t)$;
        
        Compute gradient: $\hat{\nabla}_t \gets \frac{\nabla_\theta \pi_\theta(a_t | s_t)}{\pi_\theta(a_t | s_t)}$
        
        Compute worst-case transition: $\hat{p}_{s,a} \gets \arg\min_{p\in\ambset_{s,a}} p\tr g(\xi|s'), \forall s'\in\states$
        
        Sample next state: $s_{t+1}\sim\hat{p}_{s,a}(\cdot|s_t,a_t)$;
           
        Record sample: $\xi \gets \xi + [s_t,a_t,c_t,d_t,\hat{\nabla}_t]$
        }
    
    $g\gets 0, h\gets 0$;
    
    \tcc{Loop backward and update parameters}
    \For{$t\gets T,T-1,T-2,\ldots,0$}{
        
        $g \gets \xi_{c_t} + \gamma \cdot g$;
        
        $h \gets \xi_{d_t} + \gamma \cdot h$;
        
        $\theta \gets \theta + \zeta_2(k) (g+\lambda h) \xi_{\hat{\nabla}_t}$ \tcp*{$\theta$ update}
        
        $\lambda \gets \lambda +  \zeta_1(k)(h-d_0)$ \tcp*{$\lambda$ update}
        }
	}
	\Return $\theta$;
	\caption{Robust-Constrained Policy Gradient Algorithm}    \label{alg:rcpg}
\end{algorithm*}
\Cref{alg:rcpg} presents a robust constrained policy gradient algorithm based on the gradient update rules derived above in \cref{subsec:grad_update}. This algorithm proceeds in an episodic way and update parameters based on the Monte-Carlo estimates of $g$ and $h$. Line 9 of the algorithm requires the nominal transition probability $\bar{p}_{s,a}$ and the ambiguity set $\ambset_{s,a}$ which can be some parameterized estimates. The step size schedules satisfy the standard conditions for stochastic approximation algorithms ensuring that $\theta$ update is on the fastest time-scale $\zeta_2(k)$ and the $\lambda$ update is on a slower time-scale $\zeta_1(k)$. This results in a two time-scale stochastic approximation algorithm and the convergence of it to a (local) saddle point can be shown following standard proof techniques~\citep{Borkar2009}.
\begin{figure}[b]
\centering
\includegraphics[width=0.8\textwidth]{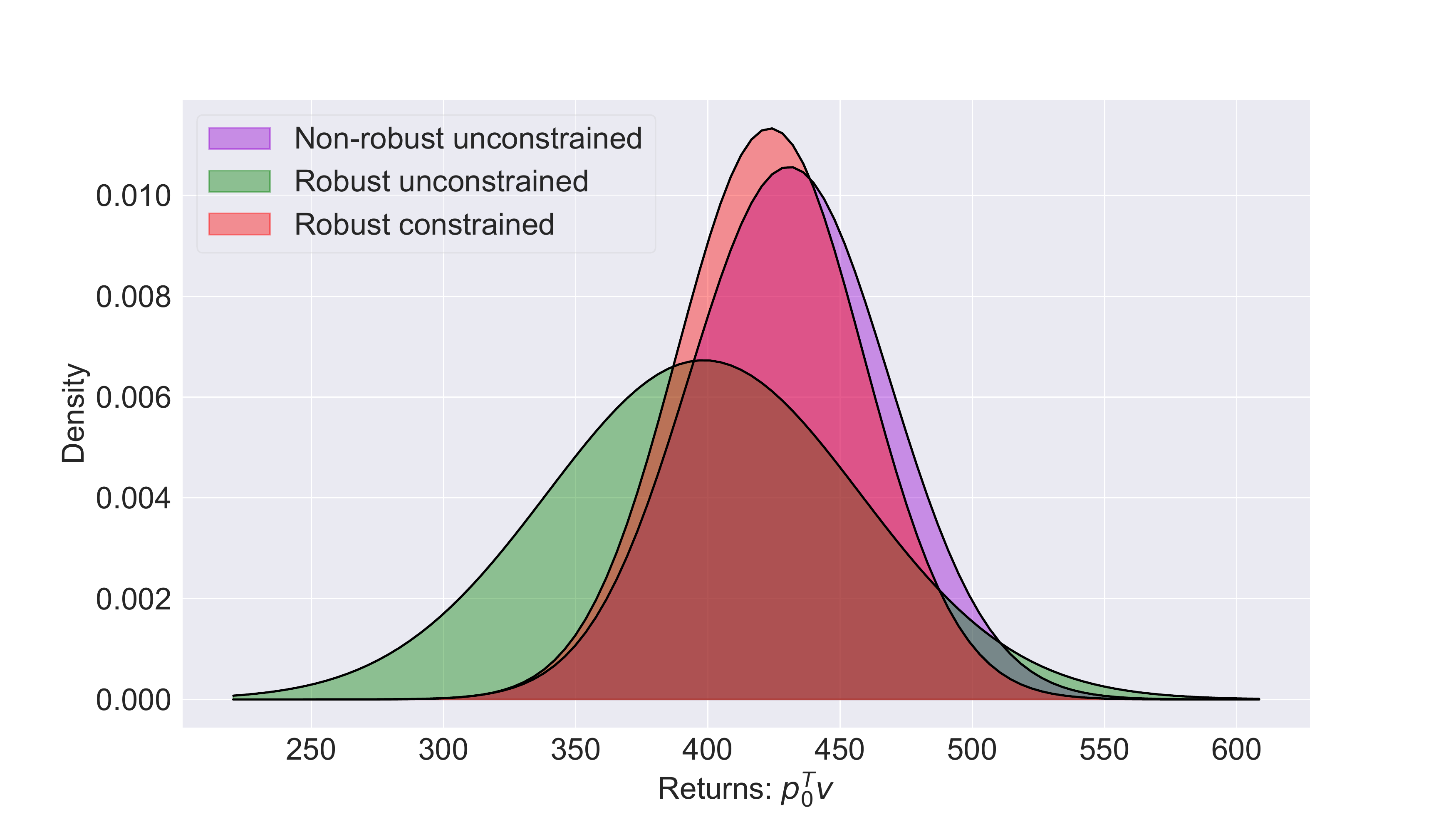}
\caption{Return distributions for the inventory management problem.} \label{fig:inventory_returns}
\end{figure}
\section{Empirical Study} \label{sec:experiment}
In this section, we empirically study the performance of our policy gradient algorithm on several problem domains. All the experiments are run with confidence parameter $\delta=0.9$, discount factor $\gamma=0.99$, and $n_{s,a}=100$ number of samples drawn for each state-action from the underlying true transition distribution $p\opt_{s,a}$. For domains emitting reward signals instead of costs, we simply consider their negative magnitude keeping our cost-based formulation intact. We implement several combinations of different settings: i) non-robust unconstrained, ii) robust-unconstrained and iii) robust-constrained. We note that the non-robust unconstrained setting is the general policy gradient algorithm~\citep{sutton1998}. The robust-unconstrained version deals only with robustness without any constraint cost or associated budget. The robust constrained version deals with both constraints and robustness as described in \cref{alg:rcpg}.

\paragraph{Inventory Management} We evaluate the policy gradient method on the classic inventory management problem~\citep{Behzadian2019,Puterman2005,Zipkin200}. The state space is discrete and is represented by the level of inventory. The goal is to order products from a supplier in order to meet demands. Demand for a product is random and comes from a normal distribution. \Cref{fig:inventory_returns} shows the return distributions for the inventory management problem. Here the non-robust unconstrained variant has slightly higher expected return compared to the robust counterparts. This is expected because the robust version particularly deals with the worst-case situation. However, the unconstrained variant is not guaranteed to enforce any constraint. The plot also shows that the robust-constrained version outperforms the robust-unconstrained variant with a higher expected return. 

\section{Conclusion} \label{sec:conclusion}
In this paper, we studied the problem of MDPs under constraints, and model uncertainties. We proposed to merge together the concepts of constrained MDPs and robust MDPs, leading to the concept of robust constrained MDPs (RCMDPs). Indeed, by doing so, one can take advantage of the safety guarantees given by the CMDP formulation, as well as the robustness guarantees w.r.t. model uncertainties, given by the RMDP formulation. We then proposed a robust soft-constrained Lagrange-based solution to the RCMDP problem, and a corresponding policy gradient algorithm. Next work will focus on extending the proposed approach to continuous domains, and validate the performance of this RCMDP formulation on more safety critical examples, e.g., robotics test-beds. 
%\newpage
\bibliographystyle{icml2019}
\bibliography{refs}

%\newpage
%\appendix
%\onecolumn

%\section{Add additional stuff}

\end{document}